\documentclass[runningheads]{llncs}
\usepackage{graphicx}
\usepackage{epstopdf}
\usepackage{cite}
\usepackage{subfigure}

\usepackage{bm}
\usepackage{amsmath,amssymb} 
\usepackage{color}

\begin{document}
\title{Quaternion Convolutional Neural Networks}

\titlerunning{Quaternion Convolutional Neural Networks}
%
\author{Xuanyu Zhu\inst{1}\thanks{Equal contribution} \and
Yi Xu\inst{1}$^\star$ \and
Hongteng Xu\inst{2,3}$^\star$ \and
Changjian Chen\inst{1}}
%
\authorrunning{Published as a conference paper at ECCV 2018}
%

%
\institute{
Shanghai Jiao Tong University, Shanghai, China\\
\email{$\{$others\_sing,xuyi$\}$@sjtu.edu.cn}, \email{ccj1988@gmail.com}
\and Infinia ML, Inc., \and Duke University, Durham, NC, USA\\
\email{hongteng.xu@duke.edu}
}

\maketitle              
\begin{abstract}
Neural networks in the real domain have been studied for a long time and achieved promising results in many vision tasks for recent years.
However, the extensions of the neural network models in other number fields and their potential applications are not fully-investigated yet.
Focusing on color images, which can be naturally represented as quaternion matrices, we propose a quaternion convolutional neural network (QCNN) model to obtain more representative features.
In particular, we re-design the basic modules like convolution layer and fully-connected layer in the quaternion domain, which can be used to establish fully-quaternion convolutional neural networks.
Moreover, these modules are compatible with almost all deep learning techniques and can be plugged into traditional CNNs easily.
We test our QCNN models in both color image classification and denoising tasks.
Experimental results show that they outperform the real-valued CNNs with same structures.
\end{abstract}

\section{Introduction}
\begin{figure}[t]
\centering
\subfigure[Real-valued CNN]{
\includegraphics[height=4.2cm]{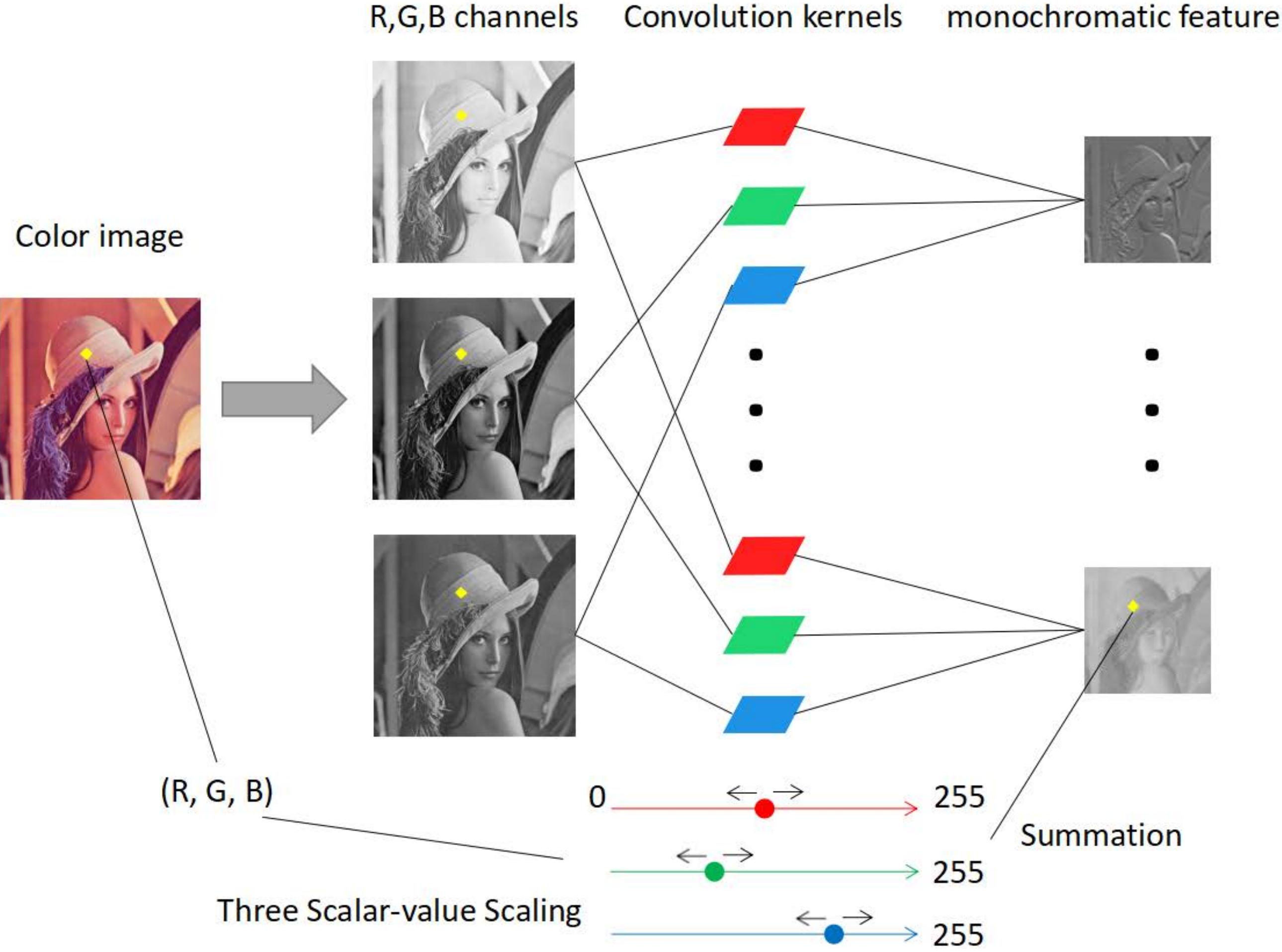}\label{fig1:cnn}
}
\subfigure[Quaternion CNN]{
\includegraphics[height=4.2cm]{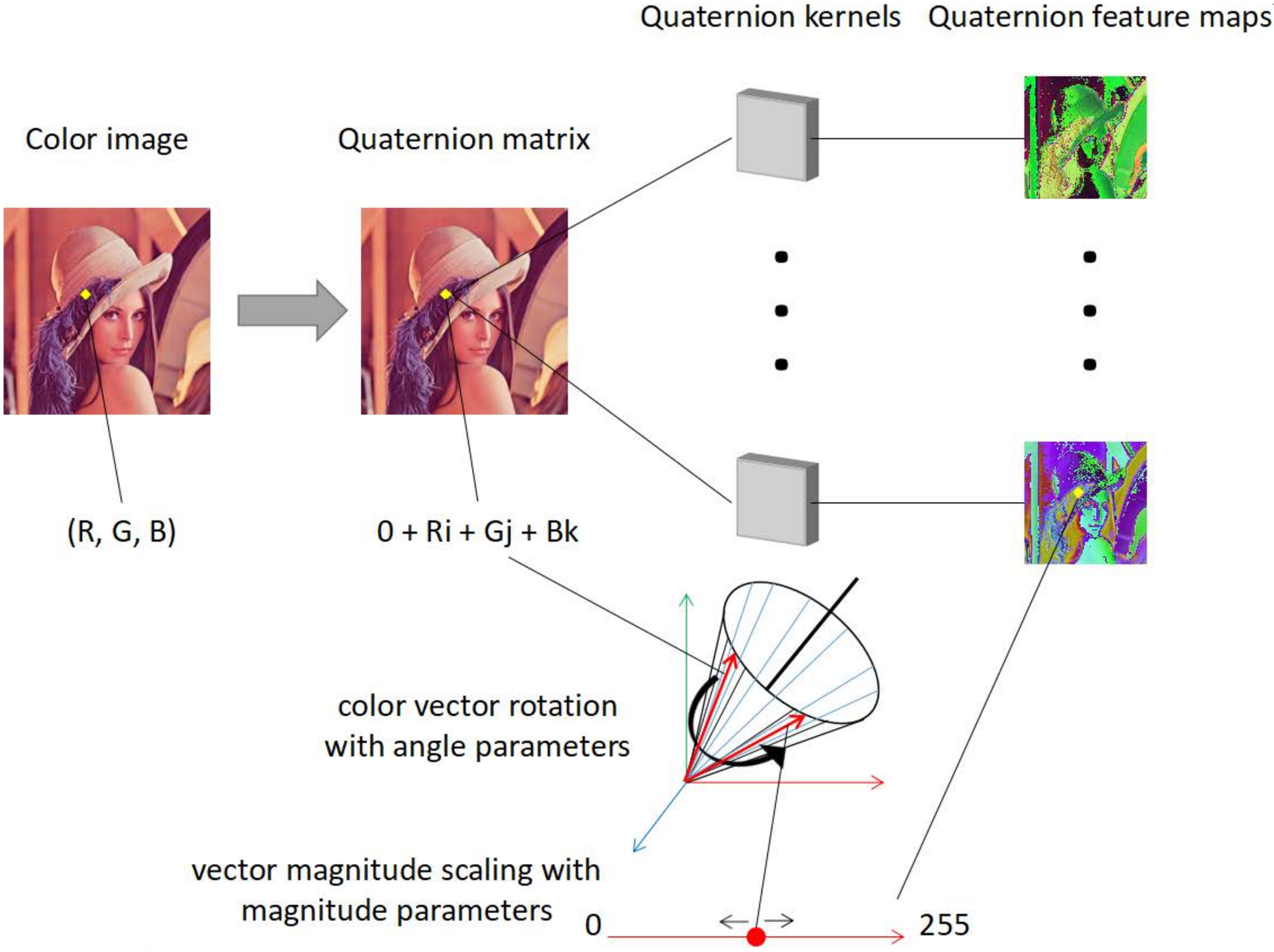}\label{fig1.qcnn}
}
\caption{Illustration of the difference between CNN and QCNN on convolution layers.}
\label{fig:Comapre}
\end{figure}

As a powerful feature representation method, convolutional neural networks (CNNs) have been widely applied in the field of computer vision.
Since the success of AlexNet~\cite{krizhevsky2012imagenet}, many novel CNNs have been proposed, $e.g.$, VGG~\cite{simonyan2014very}, ResNet~\cite{he2016deep}, and DenseNet~\cite{huang2017densely}, etc., which achieved state-of-the-art performance in almost all vision tasks~\cite{He2017Mask,cao2017realtime,long2015fully}.
One key module of CNN model is the convolution layer, which extracts features from high-dimensional structural data efficiently by a set of convolution kernels.
When dealing with multi-channel inputs ($e.g.$, color images), the convolution kernels merges these channels by summing up the convolution results and output one single channel per kernel accordingly, as Fig.~\ref{fig1:cnn} shows.

Although such a processing strategy performs well in many practical situations, it congenitally suffers from some drawbacks in color image processing tasks.
Firstly, for each kernel it just sums up the outputs corresponding to different channels and ignores the complicated interrelationship between them.
Accordingly, we may lose important structural information of color and obtain non-optimal representation of color image~\cite{xu2015vector}.
Secondly, simply summing up the outputs gives too many degrees of freedom to the learning of convolution kernels, and thus we may have a high risk of over-fitting even if imposing heavy regularization terms.
How to overcome these two challenges is still not fully-investigated.

Focusing on the problems mentioned above, we propose a novel quaternion convolutional neural network (QCNN) model, which represents color image in the  quaternion domain.
Fig.~\ref{fig:Comapre} illustrates the scheme of QCNN model.
In particular, each color pixel in a color image ($i.e.$, the yellow dot in Fig.~\ref{fig:Comapre}) is represented as a quaternion, and accordingly, the image is represented as a quaternion matrix rather than three independent real-valued matrices.
Taking the quaternion matrix as the input of our network, we design a series of basic modules, $e.g.$, quaternion convolution layer, quaternion fully-connected layer.
\textcolor{black}{While the traditional real-valued convolution is only capable to enforce scaling transformation on the input, specifically, the quaternion convolution achieves the scaling and the rotation of input in the color space, which provides us with more structural representation of color information.}
Based on these modules, we can establish fully-quaternion CNNs to represent color images in a more effective way.
Moreover, we study the relationship between our QCNN model and existing real-valued CNNs and find a compatible way to combine them together in a same algorithmic framework.

Essentially, our QCNN imposes an implicit regularizer on the architecture of network, which ensures that the representations of color image
under the guidance of quaternion operations.
Such a strategy considers more complicated relationships across different channels while suppress the degrees of freedom of model's parameters during training.
As a result, using quaternion CNNs, we can achieve better learning results with fewer parameters compared with real-valued CNNs.
Additionally, a color image is represented as a quaternion matrix in our QCNN, so that we can transform a color pixel throughout the color space using independent and physically-meaningful parameters \textcolor{black}{($i.e.$, the magnitude and the angle on the color cone shown in Fig.~\ref{fig1.qcnn})}, which enhances the interpretability of the model.
As Fig.~\ref{fig:Comapre} shows, our QCNN preserves more color information than real-valued CNN, which is suitable for color image processing, especially for low-level color feature extraction.
\textcolor{black}{Experimental results show that our QCNN model provides benefits for both high-level vision task ($i.e.$, color image classification) and low-level vision task ($i.e.$, color image denoising), which outperforms its competitors.}

\section{Related works}
\subsection{Quaternion-based color image processing}
Quaternion is a kind of hyper complex numbers, which is first described by Hamilton in 1843 and interpreted as points in three-dimensional space.
Mathematically, a quaternion $\hat{q}$ in the quaternion domain $\mathbb{H}$, $i.e.$, $q\in\mathbb{H}$, can be represented as $\hat{q} = q_0 + q_1 i + q_2 j + q_3 k$,
where $q_l\in\mathbb{R}$ for $l=0, 1, 2, 3$, and the imaginary units $i$, $j$, $k$ obey the quaternion rules that $i^2 = j^2 = k^2 = ijk = -1$.

Accordingly, a $N$-dimensional quaternion vector can be denoted as $\hat{\bm{q}}=[\hat{q}_1,...,\hat{q}_N]^{\top}\in\mathbb{H}^N$.
Similar to real numbers, we can define a series of operations for quaternions:
\begin{itemize}
\item \textbf{Addition:} $\hat{p}+\hat{q}=(p_0+q_0)+(p_1+q_1)i+(p_2+q_2)j+(p_3+q_3)k$.
\item \textbf{Scalar multiplication:} $\lambda\hat{q}=\lambda q_0 +\lambda q_1 i +\lambda q_2 j +\lambda q_3 k$.
\item \textbf{Element multiplication:}
\begin{eqnarray*}
\begin{aligned}
\hat{p}\hat{q}=&(p_0q_0-p_1q_1-p_2q_2-p_3q_3)+(p_0q_1+p_1q_0+p_2q_3-p_3q_2)i\\
&+(p_0q_2-p_1q_3+p_2q_0+p_3q_1)j+(p_0q_3+p_1q_2-p_2q_1+p_3q_0)k.
\end{aligned}
\end{eqnarray*}
\item \textbf{Conjugation:} $\hat{q}^{*}=q_0 - q_1 i - q_2 j - q_3 k$.
\end{itemize}

These quaternion operations can be used to represent rotations in a three-dimensional space.
Suppose that we rotate a 3D vector $\bm{q}=[q_1\ q_2\ q_3]^{\top}$ to get a new vector $\bm{p}=[p_1\ p_2\ p_3]^{\top}$, with an angle $\theta$ and along a rotation axis $\bm{w}=[w_1\ w_2\ w_3]^{\top}$, $w_1^2+w_2^2+w_3^2=1$.
Such a rotation is equivalent to the following quaternion operation:
\begin{eqnarray}\label{eq:rotate}
\begin{aligned}
\hat{p}=\hat{w}\hat{q}\hat{w}^*,
\end{aligned}
\end{eqnarray}
where $\hat{q}=0+q_1i+q_2j+q_3k$ and $\hat{p}=0+p_1i+p_2j+p_3k$ are pure quaternion representations of these two vectors, and
\begin{eqnarray}\label{eq:rotate_quaternion}
\begin{aligned}
\hat{w}=\cos\frac{\theta}{2}+\sin\frac{\theta}{2}(w_1i+w_2j+w_3k).
\end{aligned}
\end{eqnarray}

Since its convenience in representing rotations of 3-D vectors, quaternion is widely used in mechanics and physics~\cite{girard1984quaternion}.
In recent years, the theory of quaternion-based harmonic analysis has been well developed and many algorithms have been proposed, $e.g.$, quaternion Fourier transform (QFT)~\cite{Sangwine1996Fourier}, quaternion wavelet transform (QWT)~\cite{Xu2010QWT,bayro2006theory}, and quaternion Kalman filter~\cite{Yun2006Design,bayro2000motor}.
Most of these algorithms have been proven to work better for 3D objects than real-valued ones.
In the field of computer vision and image processing, quaternion-based methods also show its potentials in many tasks.
The advantages of quaternion wavelet transform~\cite{Jones2003Color,bayro2006theory}, quaternion principal component analysis~\cite{zeng2016color} and other quaternion color image processing techniques~\cite{Yu2013Quaternion} have been proven to extract more representative features for color images and achieved encouraging results in high-level vision tasks like color image classification.
In low-level vision tasks like image denoising and super-resolution, the quaternion-based methods~\cite{gai2016sparse,yu2014single} preserve more interrelationship information across different channels, and thus, can restore images with higher quality.
Recently, a quaternion-based neural network is also put forward and used for classification tasks~\cite{nitta1995quaternary,shang2014quaternion,bayro2001geometric}.
However, how to design a quaternion CNN is still an open problem.

\subsection{Real-valued CNNs and their extensions}
Convolutional neural network is one of the most successful models in many vision tasks.
Since the success of LeNet~\cite{lecun1989backpropagation} in digit recognition, great progresses have been made.
AlexNet~\cite{krizhevsky2012imagenet} is the first deep CNN that greatly outperforms all past models in image classification task.
Then, a number of models with deep and complicated structures are proposed, such as VGG~\cite{simonyan2014very} and ResNet~\cite{he2016deep}, which achieve incredible success in ILSVRC~\cite{deng2009imagenet}.
Recently, the CNN models are also introduced for low-level vision tasks.
For example, SRCNN~\cite{dong2014learning} applies convolutional neural networks to image super-resolution and outperforms classical methods.
For other tasks like denoising~\cite{mao2016image} and inpainting~\cite{xie2012image}, CNNs also achieve encouraging results.

Some efforts have been made to extend real-valued neural networks to other number fields.
Complex-valued neural networks have been built and proved to have advantage on generalization ability~\cite{Hirose2012Generalization} and can be more easily optimized~\cite{Nitta2002On}.
Audio signals can be naturally represented as complex numbers, so the complex CNNs are more suitable for such a kind of tasks than real-valued CNNs.
It has been proven that deep complex networks can obtain competitive results with real-valued models on audio-related tasks~\cite{trabelsi2017deep}.
In~\cite{gaudet2017deep}, a deep quaternion network is proposed.
However, its convolution simply replaces the real multiplications with quaternion ones, and its quaternion kernel is not further parameterized.
Our proposed quaternion convolution, however, is physically-meaningful for color image processing tasks.

\section{Proposed Quaternion CNNs}
\subsection{Quaternion convolution layers}\label{ssec:qcl}
Focusing on color image representation, our quaternion CNN treats a color image as a 2D pure quaternion matrix, denoted as $\widehat{\bm{A}}=[\hat{a}_{nn'}]\in\mathbb{H}^{N\times N}$, where $N$ represents the size of the image.\footnote{Without the loss of generality, we assume that both the width and the height of image are equal to $N$ in the following content.}
In particular, the quaternion matrix $\widehat{\bm{A}}$ is
\begin{eqnarray}\label{eq:qmat}
\begin{aligned}
\widehat{\bm{A}}=\bm{0}+\bm{R}i+\bm{G}j+\bm{B}k,
\end{aligned}
\end{eqnarray}
where $\bm{R}$, $\bm{G}$, $\bm{B}\in\mathbb{R}^{N\times N}$ represent red, green and blue channels, respectively.

Suppose that we have an $L\times L$ quaternion convolution kernel $\widehat{W}=[\hat{w}_{ll'}]\in\mathbb{H}^{L\times L}$.
We aim to design an effective and physically-meaningful quaternion convolution operation, denoted as ``$\circledast$'', between the input $\widehat{\bm{A}}$ and the kernel $\widehat{W}$. Specifically, this operation should $(i)$ apply rotations and scalings to color vectors in order to find the best representation in the whole color space; $(ii)$ play the same role as real-valued convolution when processing grayscale images.
\textcolor{black}{To achieve this aim, we take advantage of the rotational nature of quaternion shown in (\ref{eq:rotate},\ref{eq:rotate_quaternion}) and propose a quaternion convolution in a particular form.
Specifically, we set the element of the quaternion convolution kernel as}
\begin{eqnarray}\label{eq:qkernel}
\begin{aligned}
\hat{w}_{ll'}=s_{ll'}(\cos\frac{\theta_{ll'}}{2}+\sin\frac{\theta_{ll'}}{2}\mu),
\end{aligned}
\end{eqnarray}
where $\theta_{ll'}\in [-\pi ,\pi]$ and $s_{ll'}\in\mathbb{R}$. $\mu$ is the gray axis with unit length($i.e.$, $\frac{\sqrt{3}}{3}(i+j+k)$). As shown in Eq.~\ref{eq:rotate_quaternion}, we want a unit quaternion to perform rotation.
Accordingly, the quaternion convolution is defined as
\begin{eqnarray}\label{eq:2conv}
\begin{aligned}
\widehat{\bm{A}}\circledast\widehat{W}=\widehat{\bm{F}}=[\hat{f}_{kk'}]\in\mathbb{H}^{(N-L+1)\times (N-L+1)},
\end{aligned}
\end{eqnarray}
where
\begin{eqnarray}\label{eq:3conv}
\begin{aligned}
\hat{f}_{kk'}=\sideset{}{_{l=1}^L}\sum \sideset{}{_{l'=1}^L}\sum \frac{1}{s_{ll'}}\hat{w}_{ll'}\hat{a}_{(k+l)(k'+l')}\hat{w}_{ll'}^*.
\end{aligned}
\end{eqnarray}
The collection of all such convolution kernels formulates the proposed quaternion convolution layer.

Different from real-valued convolution operation, whose elementary operation is the multiplication between real numbers, the elementary operation of quaternion convolution in (\ref{eq:3conv}) actually applies a series of rotations and scalings to the quaternions $\hat{a}_{nn'}$'s in each patch.
The rotation axis is set as $(\frac{\sqrt{3}}{3},\frac{\sqrt{3}}{3},\frac{\sqrt{3}}{3})$ ($i.e.$, grayscale axis in color space) for all operations, while the rotation angle and the scaling factor are specified for each operation by $\theta_{ll'}$ and $s_{ll'}$, respectively.

The advantage of such a definition is interpretable.
As shown in Fig.~\ref{fig1:cnn}, the convolution in traditional CNNs operates triple scaling transforms to each pixel independently to walk through three color axes and it needs to find the best representation in the whole color space accordingly.
For our QCNNs, one pixel is a quaternion or a 3D vector in color space, but the proposed convolution find its best representation in a small part of the color space because we restrict the convolution to apply only a rotate and a scaling transform.
Such a convolution actually impose implicit regularizers on the model, such that we can suppress the risk of over-fitting brought by too many degrees of freedom to the learning of kernels.
Additionally, in real-valued CNNs, the input layer transfers 3-channel images to single-channel feature maps, ignoring the interrelationship among channels, which causes information loss.
Although the loss can be recovered with multiple different filters, the recovery requires redundant iterations, and there's no guarantee that the loss can be recovered perfectly.
In QCNNs, the convolution causes no order reduction in the input layer, thus the information of interrelationship among channels can be fully conserved.

Although our convolution operation is designed for color image, it can be applied to grayscale image as well.
For grayscale images, they can be seen as color images whose channels are the same.
Because all the corresponding color vectors are parallel to the gray axis, the rotate transform equals to identical transformation, thus the quaternion convolution performs the same function as real-valued convolution.
From this viewpoint, real-valued convolution is a special case of quaternion convolution for grayscale image.

According to the rule of quaternion computations, if we represent each $\hat{a}_{nn'}$ as a 3D vector $\bm{a}_{nn'}=[a_1\ a_2\ a_3]^{\top}$, then the operation in~(\ref{eq:3conv}) can be represented by a set of matrix multiplications:
\begin{eqnarray}\label{eq:4conv}
\begin{aligned}
\bm{f}_{kk'}=\sideset{}{_{l=1}^L}\sum \sideset{}{_{l'=1}^L}\sum s_{ll'}\left(
  \begin{array}{ccc}
    f_1 & f_2 & f_3\\
    f_3 & f_1 & f_2\\
    f_2 & f_3 & f_1\\
  \end{array}
\right)\bm{a}_{(k+l)(k'+l')},
\end{aligned}
\end{eqnarray}
where $\bm{f}_{kk'}$ is a vectorized representation of quaternion $\hat{f}_{kk'}$, and
\begin{eqnarray}\label{eq:f}
\begin{aligned}
f_1 = \frac{1}{3}+\frac{2}{3}\cos\theta_{ll'},~f_2 =\frac{1}{3} -\frac{2}{3}\cos(\theta_{ll'}-\frac{\pi}{3}),~
f_3 = \frac{1}{3}-\frac{2}{3}\cos(\theta_{ll'}+\frac{\pi}{3}).
\end{aligned}
\end{eqnarray}
The detailed derivation from (\ref{eq:3conv}) to (\ref{eq:4conv}) is given in the supplementary file.
Additionally, because the inputs and outputs of quaternion convolutions are both pure quaternion matrices, quaternion convolution layers can be stacked like what we do in real-valued CNNs and most architectures of real-valued CNNs can also be used in QCNNs.
In other words, the proposed quaternion convolution is compatible with traditional real-valued convolution.

According to (\ref{eq:4conv}), we can find that a quaternion convolution layer has twice as many parameters as the real-valued convolution layer with same structure and same number of filtering kernels since an arbitrary element of quaternion convolution kernel has two trainable parameters $s$ and $\theta$.
Denote $K$ as the number of kernels, $L$ as kernel size and $C$ as the number of input channels.
A real-valued convolution layer with $K$ $L\times L\times C$ kernels has $KCL^2$ parameters, and we require $L^2N^2KC$ multiplications to process $C$ $N\times N$ feature maps.
A quaternion layer with $K$ $L\times L\times C$ kernels has $2KCL^2$ parameters: each kernel has $CL^2$ angle parameters $[\theta_{ll'c}]$ and $CL^2$ scaling parameters $[s_{ll'c}]$.
To process $C$ $N\times N\times 3$ color feature maps, we need $9L^2N^2KC$ multiplications because each output quaternion $\bm{f}_{kk'}$ requires $9L^2$ multiplications, as shown in the (\ref{eq:4conv}).
By reducing the number of the kernels and channels to $\frac{K}{\sqrt{2}}$ and $\frac{C}{\sqrt{2}}$, the number of the quaternion layer's parameters is halved and equal to that of real-valued layer.
Since the number of channels $C$ in one layer is equal to the number of kernels $K$ in the previous layer, by reducing the number of the kernels in all layers with the ratio $\frac{1}{\sqrt{2}}$, we half the number of QCNN's parameters and half the number of operations to 4.5 times as that of the real-valued CNN. Note that the matrix multiplication in the (\ref{eq:4conv}) can be optimized and parallelized when implemented by Tensorflow.
In our experiments, our QCNNs only takes about twice as much time as real-valued CNNs with same number of parameters.
According to our following experiments, such a simplification will not do harm to our QCNN model --- experimental results show that the QCNNs with comparable number of parameters to real-valued CNNs can still have superior performance.

\subsection{Quaternion fully-connected layers}
The quaternion convolution layer mentioned above preserves more interrelationship information and extracting better features than real-valued one.
However, if we had to connect it to a common fully-connected layer, that kind of information preserved would be lost.
Therefore, here we design a quaternion fully-connected layer that performs same operation as quaternion convolution layer to keep the interrelationship information between channels.
Specifically, similar to the real-valued CNNs, whose fully-connected layers can be seen as special cases of one-dimensional convolution layers with kernels having same shapes with inputs, our quaternion fully-connected layers follow the same rule.
Suppose that the input is an $N$-dimensional quaternion vector $\hat{\bm{a}}=[\hat{a}_i]\in\mathbb{H}^{N}$, for $i=1,2,3...N$.
Applying $M$ 1D quaternion filtering kernels, $i.e.$, $\hat{\bm{w}}^m=[\hat{w}_i^m]\in\mathbb{H}^M$ for $m=1,..,M$, we obtain an output $\hat{\bm{b}}=[\hat{b}_m]\in\mathbb{H}^M$ with element
\begin{eqnarray}\label{eq:1Fc}
\begin{aligned}
\hat{b}_{m}=\sideset{}{_{i=1}^N}\sum \frac{1}{s_{i}}\hat{w}_{i}^m\hat{a}_{i}\hat{w}_{i}^{m*},
\end{aligned}
\end{eqnarray}
where $s_i$ is the magnitude of $\hat{w}_{i}^m$.

Similar to our quaternion convolution layer, the computation of the proposed quaternion fully-connected layer can also be reformulated as a set of matrix multiplications, and thus, it is also compatible with real-valued CNNs.

\subsection{Typical nonlinear layers}
Pooling and activation are import layers to achieve nonlinear operations.
For our QCNN model, we extend those widely-used real-valued nonlinear layers to quaternion versions.
For average-pooling, the average operation of quaternion is same as averaging the 3 imaginary parts respectively.
For max-pooling, we can define various criterions such as magnitude or projection to gray axis to judge which element to choose.

In our experiments, we find that simply applying max-pooling to 3 imaginary parts respectively can provides us with good learning results.
Similarly, we use same activation functions with real-valued CNNs for each channel respectively in QCNNs. For ReLU, if a vector of quaternion is rotated out of valid value range in color space, e.g. negative color value for RGB channels, we reset it to the nearest point in color space.

For softmax, we split the output of the quaternion layer in to real numbers and connect them to real-valued softmax layers and train classifiers accordingly.

\subsection{Connecting with real-valued networks}\label{sec:Connect}
Using the modules mentioned above, we can establish arbitrary fully-quaternion CNNs easily.
Moreover, because of the compatibility of these modules, we can also build hybrid convolutional neural networks using both quaternion-based layers and common real-valued layers.
In particular,
\begin{itemize}
\item \textbf{Connect to real-valued convolution layer:} The feature map that a quaternion layer outputs can be split into 3 grayscale feature maps, each corresponding to one channel.
Then, we can connect each of these three maps to real-valued convolution layers independently, or concatenate them together and connect with a single real-valued convolution layers.
\item \textbf{Connect to real-valued fully-connected layer:} Similarly, we flatten the output of a quaternion layer and treat each quaternion element as 3 real numbers.
Thus, we obtain a real-valued and vectorized output which can be connected to real-valued fully-connected layer easily.
\end{itemize}

\section{Learning Quaternion CNNs}
\subsection{Weight initialization}
Proper weight initialization is essential for a network to be successfully trained.
This principle is also applicable to our QCNN model.
According to our analysis above, the scaling factor $s$ corresponds to the parameters in real-valued CNNs, which controls the magnitude of transformed vector, while the rotation angle $\theta$ is an additional parameter, which only makes the transformed vector an rotation of input vector.
Additionally, when transformed vectors are added together, though the magnitude is affected by $\theta$, its projection to gray axis is still independent of $\theta$.
Therefore, we follow the suggestion proposed in~\cite{Glorot2010Understanding} and perform normalized initialization in order to keep variance of the gradients same during training.
Specifically, for each scaling factor and each rotation factor of the $j$-th layer, $i.e.$, $s_j$ and $\theta$, and we initialize them as two uniform random variables:
\begin{eqnarray}
\begin{aligned}
s_j \sim U\left[-\frac{\sqrt{6}}{\sqrt{n_j+n_{j+1}}},\frac{\sqrt{6}}{\sqrt{n_j+n_{j+1}}}\right],\quad \theta \sim U\left[-\frac{\pi}{2},\frac{\pi}{2}\right].
\end{aligned}
\end{eqnarray}
where $U[\cdot]$ represents a uniform distribution, and $n_j$ means the dimension of the $j$-th layer's input.

\subsection{Backpropagation}\label{ssec:back}
Backpropagation is the key of training a network, which applies the chain rule to compute gradients of the parameters and updates them.
Denote $L$ as the real-valued loss function used to train our quaternion CNN model.
$\hat{p}=p_1 i+p_2 j+ p_3 k$ and $\hat{q}=q_1 i+q_2 j+q_3 k$ are two pure quaternion variables.
For the operation we perform in the QCNN, $i.e.$, $\hat{p}=\frac{1}{s} \hat{w}\hat{q}\hat{w}^{*}$, it can be equivalently represented by a set of matrix multiplications.
So is the corresponding quaternion gradient.
Particularly, we have:
\begin{eqnarray}
\begin{aligned}
\frac{\partial L}{\partial \vec q}=\frac{\partial L}{\partial \vec p}\frac{\partial \vec p}{\partial \vec q},~~
\frac{\partial L}{\partial \theta}=\frac{\partial L}{\partial \vec p}\frac{\partial \vec p}{\partial \theta},~~
\frac{\partial L}{\partial s}=\frac{\partial L}{\partial \vec p}\frac{\partial \vec p}{\partial s},
\end{aligned}
\end{eqnarray}
where $\bm{p}=[p_1,p_2,p_3]^\top$ and $\bm{q}=[q_1,q_2,q_3]^\top$ are vectors corresponding to $\hat{p}$ and $\hat{q}$.
When $\bm{p}$ and $\bm{q}$ are arbitrary elements of feature maps and filtering kernels, corresponding to $\bm{a}_{nn'}$ and $\bm{w}_{ll'}$ in (\ref{eq:4conv}), we have
\begin{eqnarray}
\begin{aligned}
&\frac{\partial \vec p}{\partial \vec q}
=s
\left(
  \begin{array}{ccc}
    f_1 & f_2 & f_3\\
    f_3 & f_1 & f_2\\
    f_2 & f_3 & f_1\\
  \end{array}
\right),~
\frac{\partial \vec p}{\partial \theta}
=s
\left(
  \begin{array}{ccc}
    f_1^{'} & f_2^{'} & f_3^{'}\\
    f_3^{'} & f_1^{'} & f_2^{'}\\
    f_2^{'} & f_3^{'} & f_1^{'}\\
  \end{array}
\right)
\left(
  \begin{array}{ccc}
    q_1\\
    q_2\\
    q_3\\
  \end{array}
\right),~
\frac{\partial \vec p}{\partial s} =
\left(
  \begin{array}{ccc}
    f_1 & f_2 & f_3\\
    f_3 & f_1 & f_2\\
    f_2 & f_3 & f_1\\
  \end{array}
\right)
\left(
  \begin{array}{ccc}
    q_1\\
    q_2\\
    q_3\\
  \end{array}
\right)
\end{aligned}
\end{eqnarray}
where $f_i$, $i=1,2,3$, is defined as (\ref{eq:f}) does.
The matrix of $f_i$'s is exactly same as that in (\ref{eq:4conv}), but the operation switches from left multiplication to right multiplication.
In other words, the backward process can be explained as a rotate transform with the same axis and a reverse angle.

\subsection{Loss and activation functions}
In neural networks, loss and activation functions must be differentiable for the gradient to generate and propagate.
For fully-quaternion CNNs, any functions which are differentiable with respect to each part of the quaternion variables also make the quaternion chain rule hold, and thus, can be used as loss (and activation) functions.
For hybrid CNNs, we select loss functions according to the category of tasks.
In classification tasks, the top of the networks are real-valued fully-connected layers, before which the quaternion inputs are flattened as section~\ref{sec:Connect} suggested, and the loss function is cross entropy loss.
In other tasks ($e.g.$, regression tasks) that the network outputs images, quaternion outputs of the top layer are regarded as the 3-channel images, and the loss function can be mean square error (MSE) or other similar functions.

\begin{table}[t]
\begin{center}
\caption{Experiment results in classification tasks}
\label{table:classification}
\begin{tabular}{lcc}
\hline\hline
Model & Dataset & Test accuracy \\
\hline
Shallow real network &  Cifar-10  & 0.7546\\
Shallow quaternion network &  Cifar-10  & \textbf{0.7778}\\ \hline
Real-valued VGG-S &  102 flowers  & 0.7308\\
Quaternion VGG-S &  102 flowers  & \textbf{0.7695}\\
Quaternion VGG-S with fewer filters &102 flowers  & \textbf{0.7603}\\
\hline\hline
\end{tabular}
\end{center}
\end{table}

\section{Experiments}
To demonstrate the superiority and the universality of our QCNN model, we test it on two typical vision tasks: color image classification and color image denoising.
These two tasks represent typical high-level and low-level vision tasks.
Compared with real-valued CNN models in these two tasks, our QCNN models show improvements on learning results consistently.
\textbf{Some typical experimental results are shown and analyzed below, and more representative results and details are given in the supplementary file.}

\subsection{Color image classification}
We have tested two QCNN architectures in our research, a shallow network for cifar-10~\cite{krizhevsky2009learning}, and a relatively deep one for 102 Oxford flowers~\cite{nilsback2008automated}.
For comparison, real-valued networks with same structure and comparable number of parameters are also trained in the same datasets.
Both quaternion and real-valued networks use a real-valued fully-connected layer with softmax function, or a softmax layer to classify the input images.
The real-valued networks use ReLU as activation functions, while the quaternion ones adapt ReLU for each imaginary part separately.
All those networks are trained with cross entropy loss.
Input data is augmented by shifting and flipping.

The proposed shallow network for cifar-10 contains 2 convolution blocks, each has 2 convolution layers and a max-pooling layer, and ends with 2 fully-connected layers.
In the experiment, each layer of real-valued CNN and QCNN are of same number of filters, so actually QCNN has more parameters.
Both models are optimized using RMSProp~\cite{hinton2012neural} with learning rate set at 0.0001, and learning rate decay set at 1e-6.
The training ends at epoch 80.

The network for 102 Oxford flowers is VGG-S~\cite{chatfield2014return}, which has 5 convolution layers, 3 pooling layers and 3 fully-connected layers.
In this experiment, a QCNN with same number of filters as real-valued one and another one with fewer filters to keep the similar number of parameters are both tested.
Models are optimized using Adam~\cite{kingma2014adam} with learning rate set at 0.0001.
The training ends at epoch 50.

\begin{figure}[ht]
\centering
\subfigure[cifar-10: training loss]{
\includegraphics[height=2.5cm,width=2.7cm]{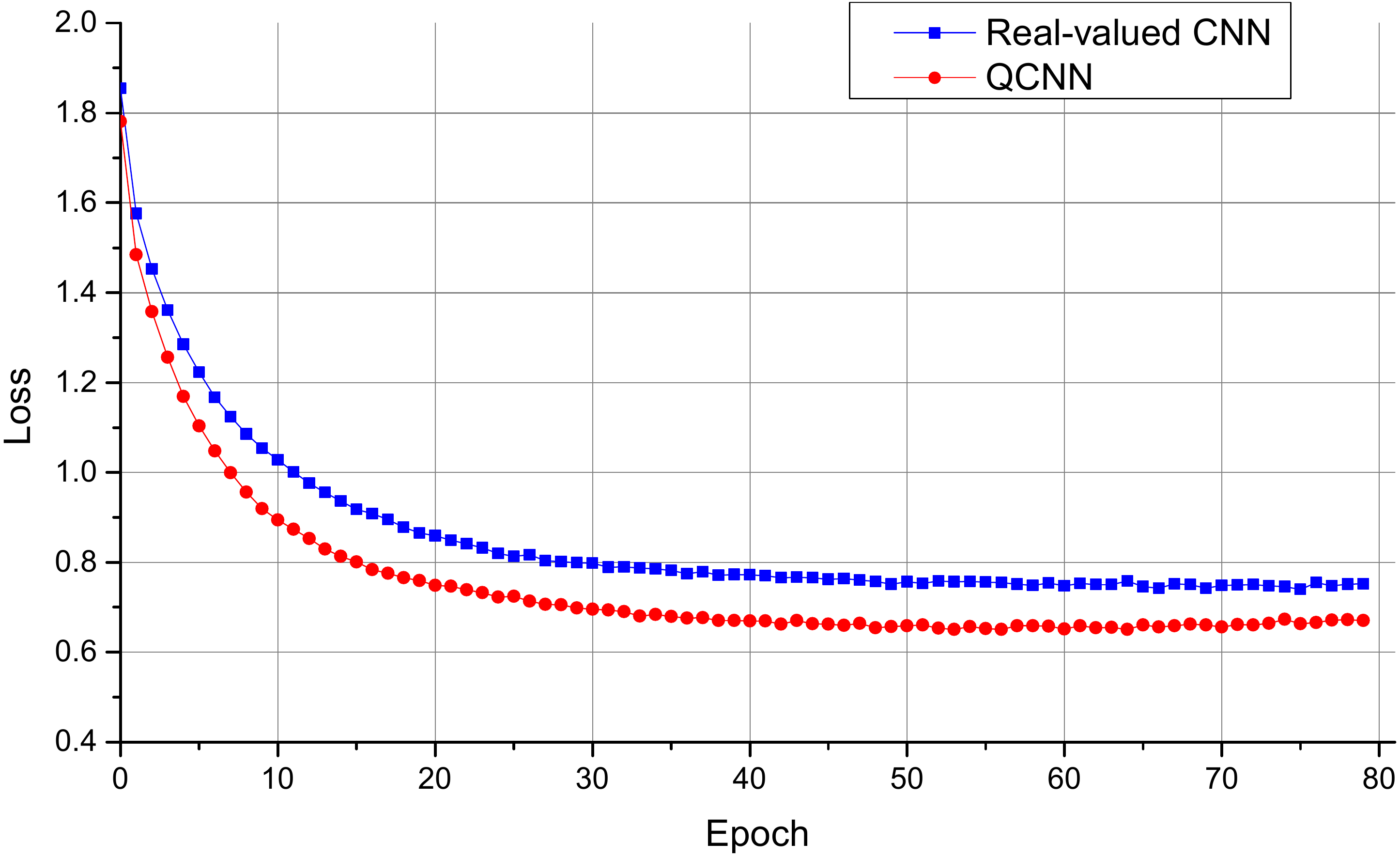}
}
\subfigure[cifar-10: classification accuracy]{
\includegraphics[height=2.5cm,width=2.7cm]{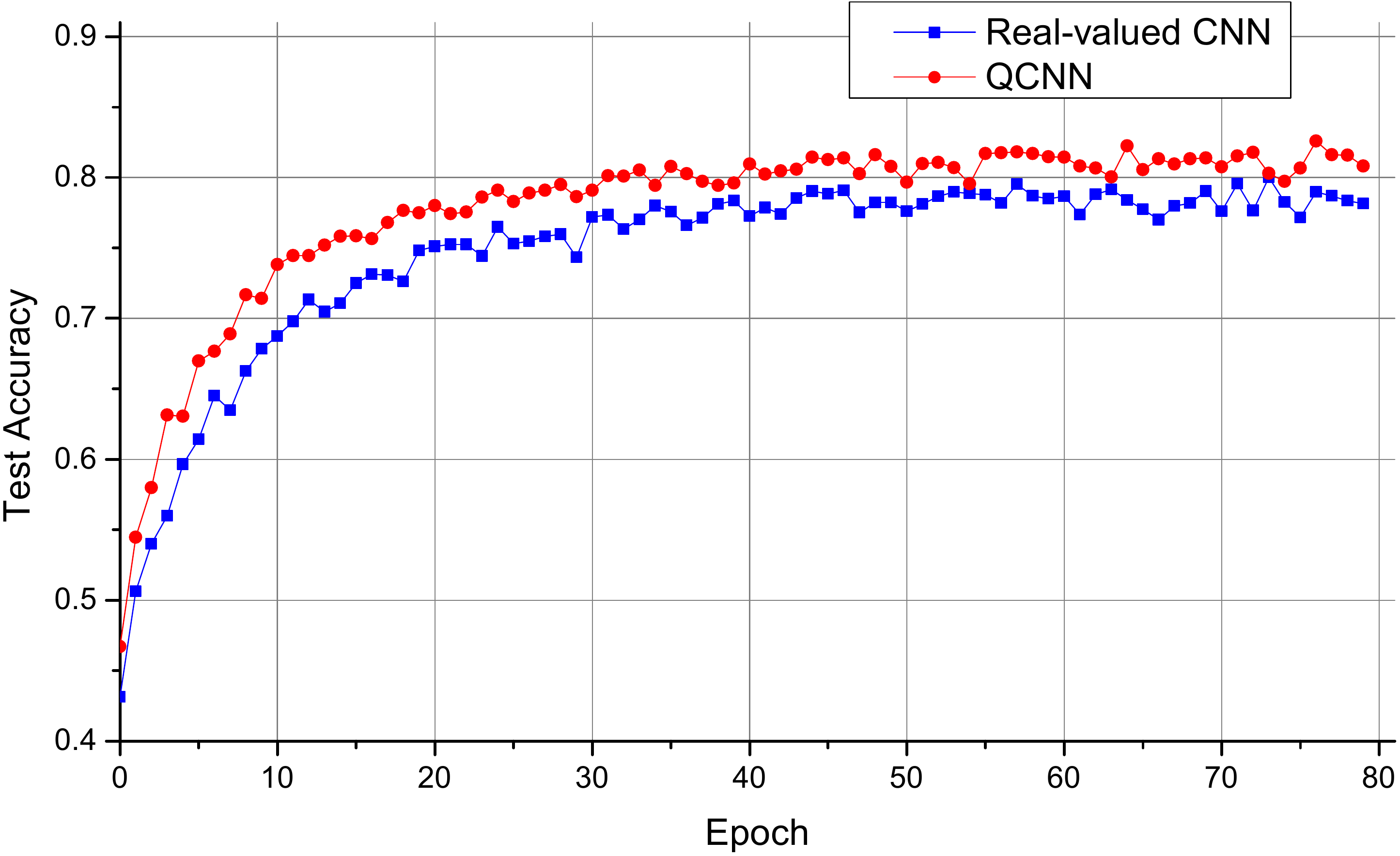}
}
\subfigure[flower: training loss]{
\includegraphics[height=2.5cm,width=2.7cm]{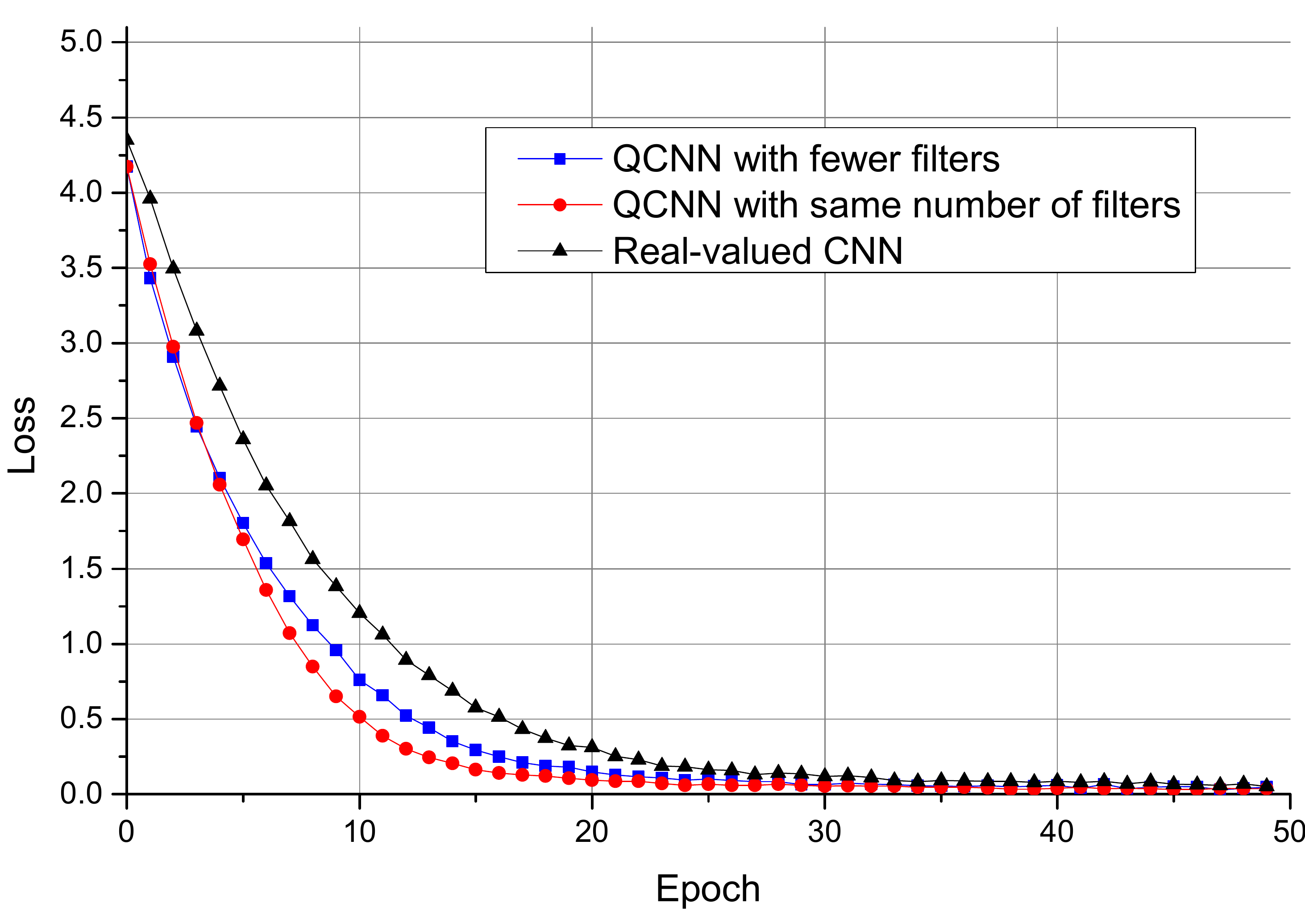}
}
\subfigure[flower: classification accuracy]{
\includegraphics[height=2.5cm,width=2.7cm]{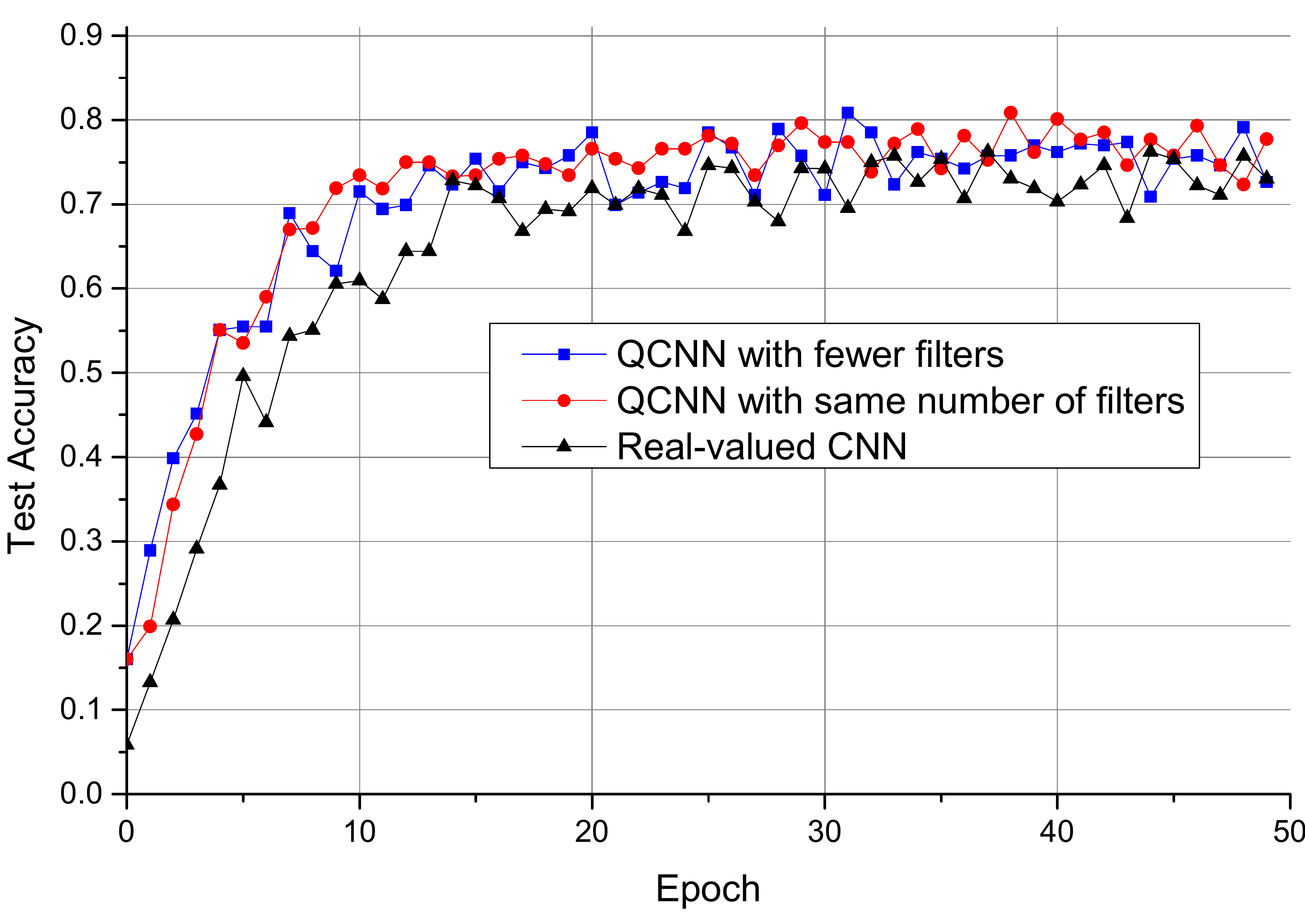}
}
\caption{(a, b) The loss and the test accuracy of the shallow networks during training on cifar-10.
(c, d) The loss and the test accuracy of the VGG-S networks during training on 102 Oxford flower data set (256 test images picked randomly from the test set for each epoch).}
\label{fig:cifar}
\end{figure}

In Fig.~\ref{fig:cifar}, we can find that the performance of our QCNNs is consistently better than that of real-valued CNNs.
For each data set, the loss function of our QCNN converges more quickly than that of real-valued CNNs in the training phase and reaches smaller loss finally.
The classification accuracy on the testing set obtained by our QCNN is also better than that of real-valued CNN even in the very beginning of training phase.
Moreover, even if we reduce the number of QCNN's parameters, the proposed QCNN model is still superior to the real-valued CNN with the same size.
These phenomena verify our claims before.
Firstly, although a QCNN can have more parameters than real-valued CNN, it can suffer less from the risk of over-fitting because of the implicit regularizers imposed by the computation of quaternions.
Secondly, the quaternion convolution achieves both the scaling and the rotation of  inputs in color space, which preserves more discriminative information for color images, and this information is beneficial for classifying color images, especially for classifying those images in which the objects have obvious color attributes ($i.e.$, the flowers in 102 Oxford flower data set).
The quantitative experimental results are given in Table~\ref{table:classification}, which further demonstrates the superiority of our model.

\begin{table}[t]
\begin{center}
\caption{Experiment results in denoising tasks}
\label{table:denoising}
\begin{tabular}{c|cc|cc}
\hline\hline
Model & Dataset & Test PSNR (dB) & Dataset & Test PSNR (dB) \\
\hline
Real-valued CNN & 102 flowers  & 30.9792          & subset of COCO  & 30.4900\\
Quaternion CNN  & 102 flowers  & \textbf{31.3176} & subset of COCO  & \textbf{30.7256}\\
\hline\hline
\end{tabular}
\end{center}
\end{table}

\subsection{Color image denoising}
Besides the high-level vision tasks like image classification, the proposed QCNN can also obtain improvements in the low-level vision tasks.
In fact, because our QCNN model can obtain more structural representation of color information, it is naturally suitable for extracting low-level features and replacing the bottom convolution layers of real-valued CNNs.
To demonstrate our claim, we test our QCNN model in color image denoising task.
Inspired from the encoder-decoder networks with symmetric skip connections for image restoration~\cite{mao2016image} and denoising autoencoders~\cite{vincent2008extracting}, a U-Net-like~\cite{ronneberger2015u} encoder-decoder structure with skip connections is used for denoising in our research.
The encoder contains two $2\times 2$ average-pooling layers, each following after two $3\times 3$ convolution layers, then two $3\times 3$ convolution layers and a fully-connected layer.
The decoder is symmetrical to the encoder, containing up-sampling and transposed convolution layers.
The layers before pooling and that after up-sampling are connected by shortcuts.
A QCNN and a real-valued CNN with this structure are both built, and the QCNN has fewer filters each layer to ensure a similar number of parameters to the real-valued CNN.
Similar to networks for classification, both networks use ReLU as activation functions except the top layer, whose activation function is ``tanh'' function.
Both networks are trained with MSE loss.

\begin{figure}[t]
\centering
\subfigure[Training loss]{
\includegraphics[height=2.5cm,width=2.7cm]{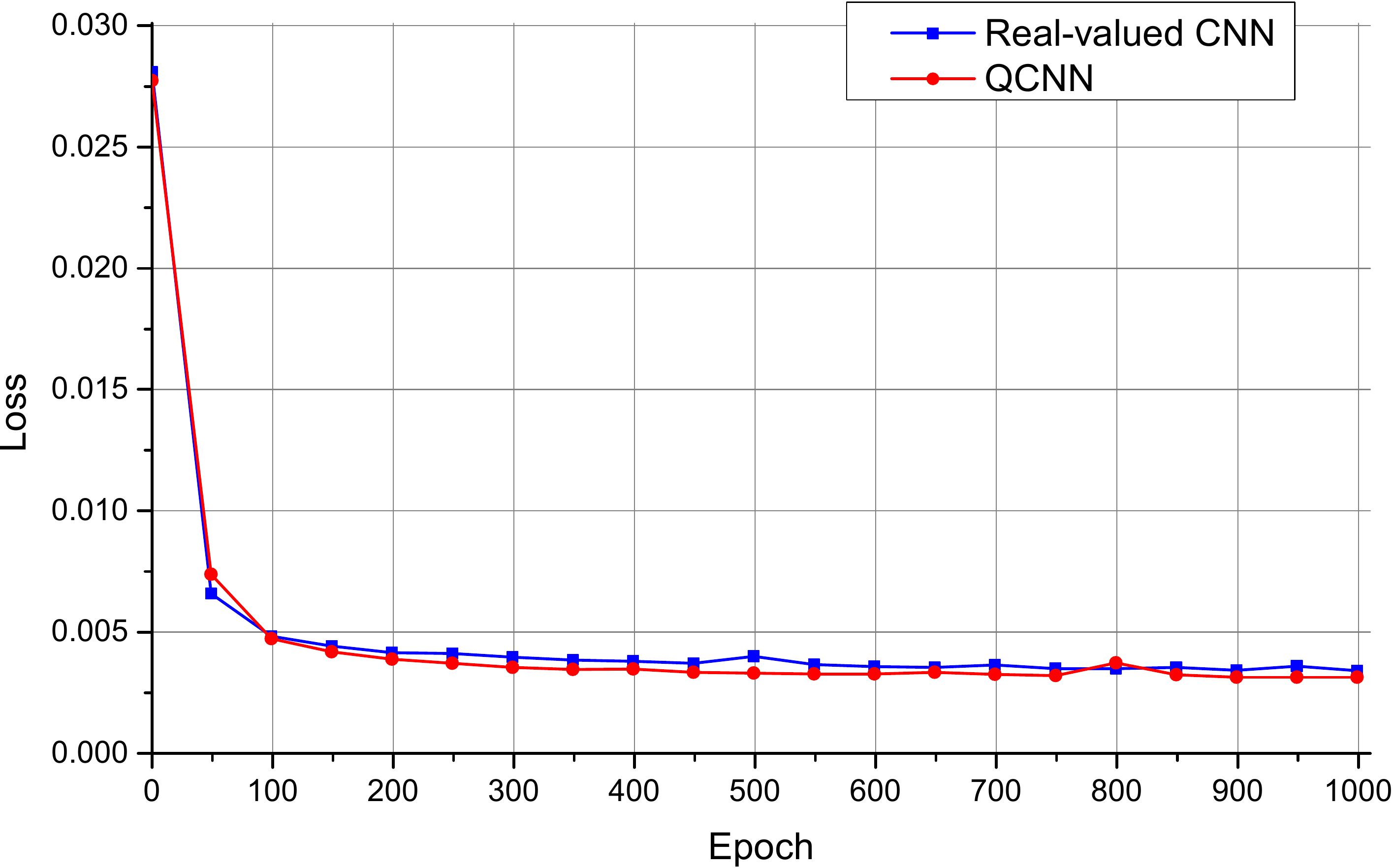}
}
\subfigure[PSNR]{
\includegraphics[height=2.5cm,width=2.7cm]{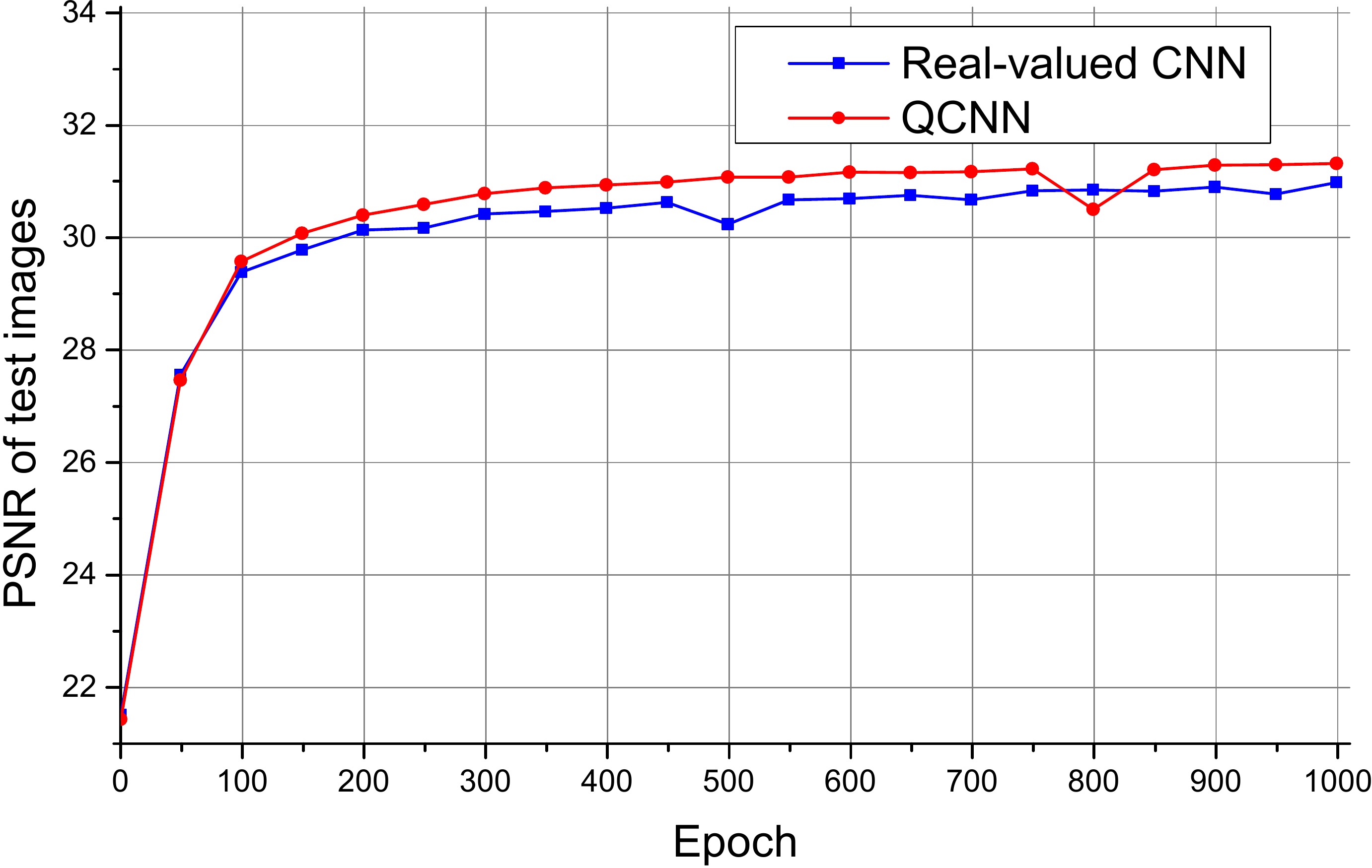}
}
\subfigure[Training loss]{
\includegraphics[height=2.5cm,width=2.7cm]{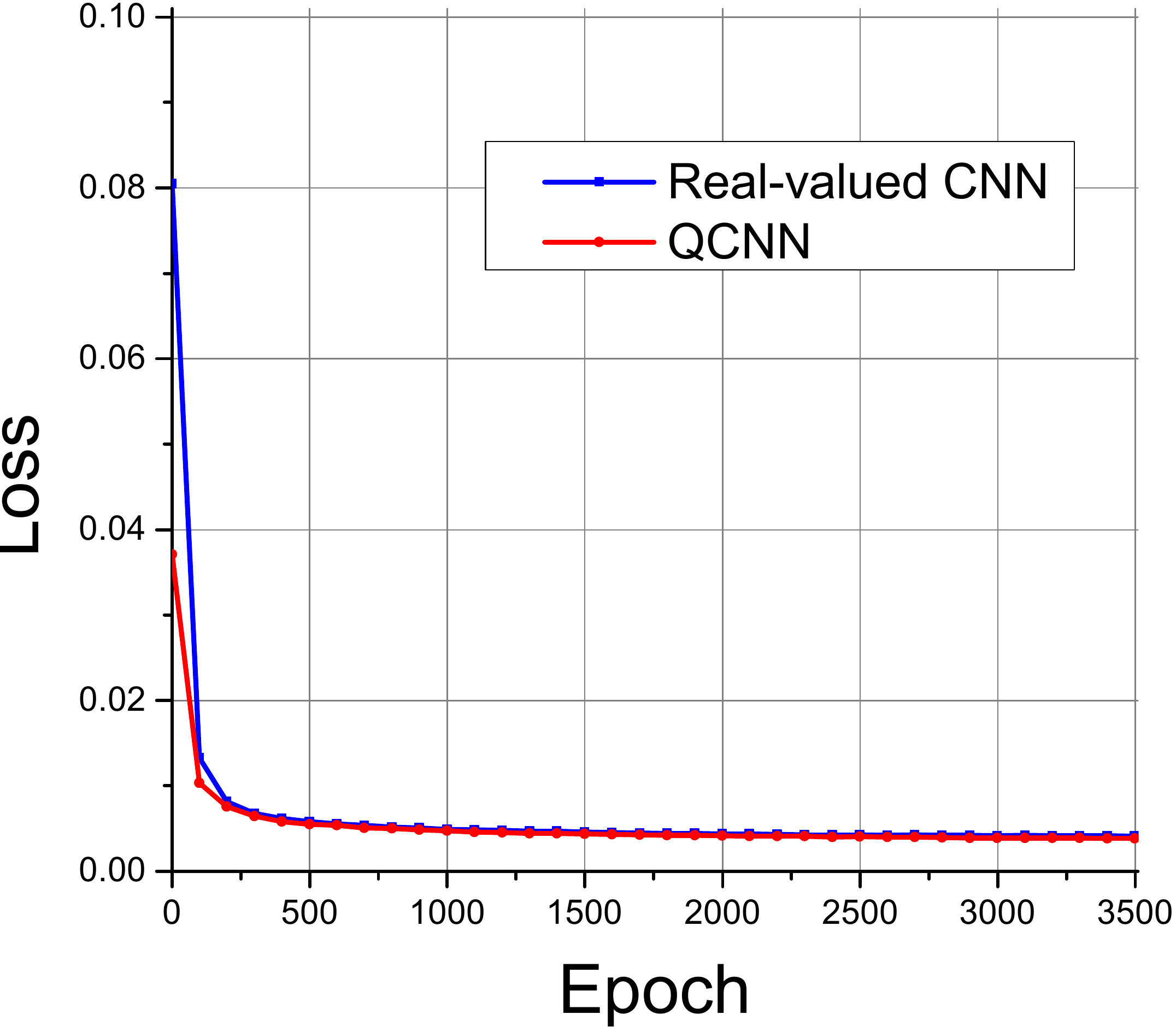}
}
\subfigure[PSNR]{
\includegraphics[height=2.5cm, width=2.7cm]{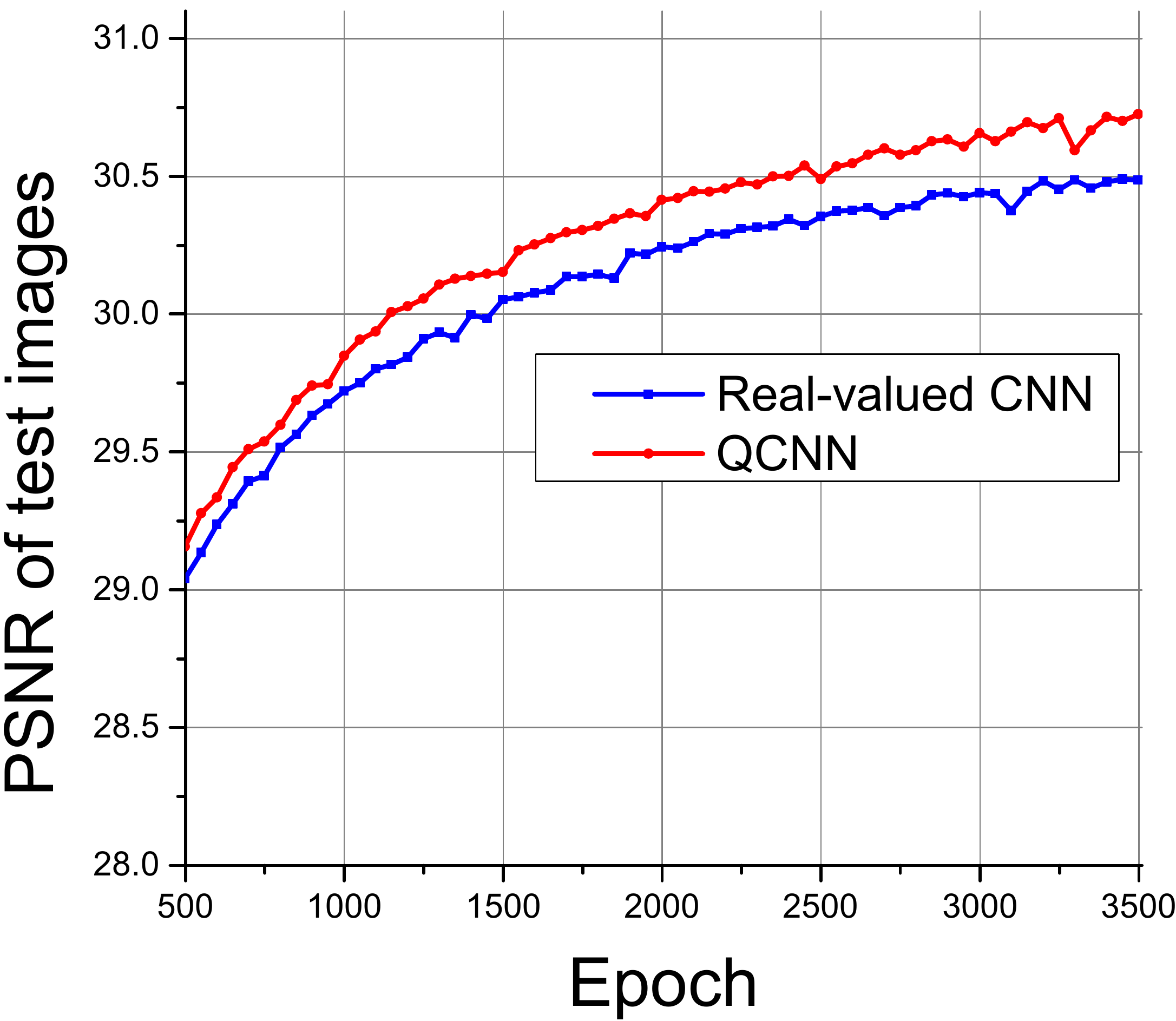}
}
\caption{(a, b) The loss and the PSNR of test images using proposed denoising networks during training on 102 Oxford flower data set. (c, d) The loss and the PSNR of test images using proposed denoising networks during training on COCO subset.}
\label{fig:COCO_train}
\end{figure}

\begin{figure}
\centering
\subfigure[Original image]{
\includegraphics[height=2cm]{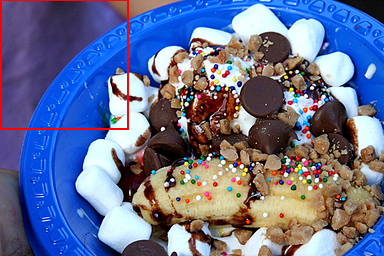}
}
\subfigure[Noisy image]{
\includegraphics[height=2cm]{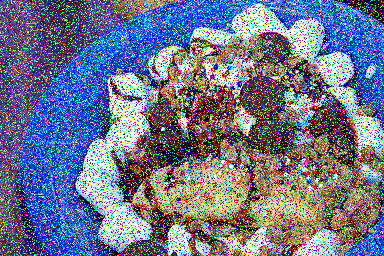}
}\\
\subfigure[Enlarged image]{
\includegraphics[height=2.9cm]{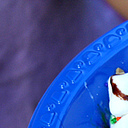}
}
\subfigure[QCNN, \textbf{25.69dB}]{
\includegraphics[height=2.9cm]{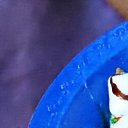}
}
\subfigure[CNN, 24.80dB]{
\includegraphics[height=2.9cm]{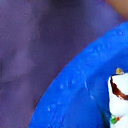}
}
\caption{Denoising experiment on a image of snacks.}
\label{fig:snack}
\end{figure}

We trained and tested these two models on two data sets: the 102 Oxford flower data set and a subset of COCO data set~\cite{lin2014microsoft}.
These two data sets are representative for our research: the flower data set is a case having colorful images, which is used to prove the superiority of our QCNN model conceptually; while the COCO subset is a more general set of natural images,  which have both colorful and colorless images and can be used to prove the performance of our model in practice.

In our experiments, both the training and the testing images are cut and resized to $128\times 128$ pixels with values normalized to $[0, 1]$.
Then a salt and pepper noise which corrupts $30\%$ of pixels and a Gaussian noise with zero mean and $0.01$ variance are added.
The inputs of networks are corrupted images, and target outputs are original images.
For both real-valued CNN and our QCNN model, the optimizer is Adam with 0.001 learning rate, and the batch size is 64 for the 102 Oxford flower data set and 32 for the COCO subset, respectively.

Table~\ref{table:denoising} shows quantitative comparisons for the real-valued CNN model and the proposed QCNN model.
We can find that our QCNN model obtains higher PSNR values consistently on both data sets.
The change of loss function and that of PSNR on testing set are given in Fig.~\ref{fig:COCO_train} for the two data sets.
Similar to the experiments in color image classification task, the loss function of our QCNN converges more quickly to a smaller value and its PSNR on testing images becomes higher than that of the real-valued CNN after 100 epochs.
Furthermore, we show a visual comparison for the denosing results of the real-valued CNN and our QCNN in Fig.~\ref{fig:snack}.
We can find that our QCNN preserves more detailed structures in the image ($e.g.$, the pattern on the plate) than the real-valued CNN does.
Suffering from information loss during feature encoding, real-valued CNNs cannot perfectly preserve the details of color images, especially when the structure presents sharp color variations.
Our QCNN, on the contrary, can avoid this information loss and learn more texture features even in bottom layers, so it outputs images of higher fidelity.
High-resolution visual comparisons can be found in the supplementary file.


\begin{figure}[t]
\centering
\subfigure[Colorful images]{
\includegraphics[height=2.7cm]{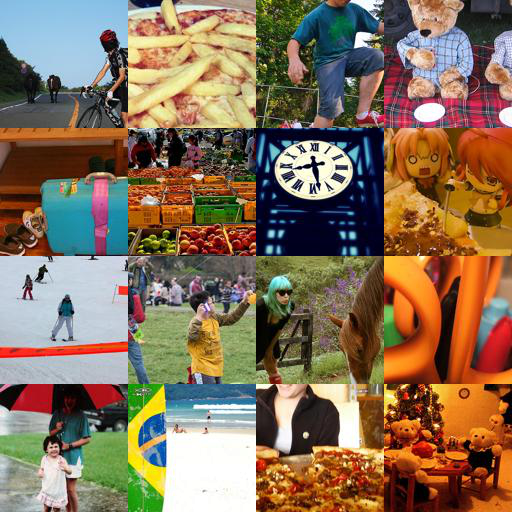}
}
\subfigure[Colorless images]{
\includegraphics[height=2.7cm]{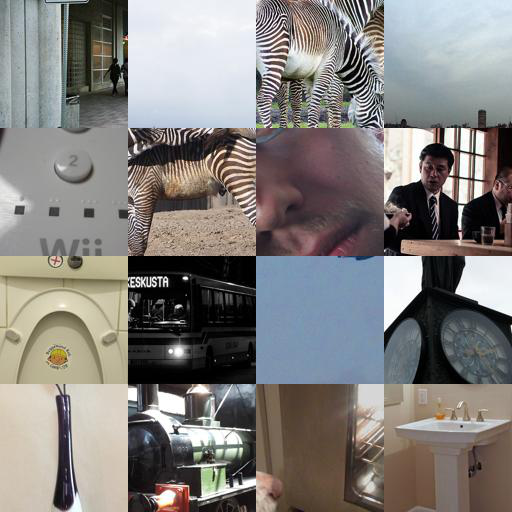}
}
\subfigure[$S$ v.s. $D$]{
\includegraphics[height=2.7cm]{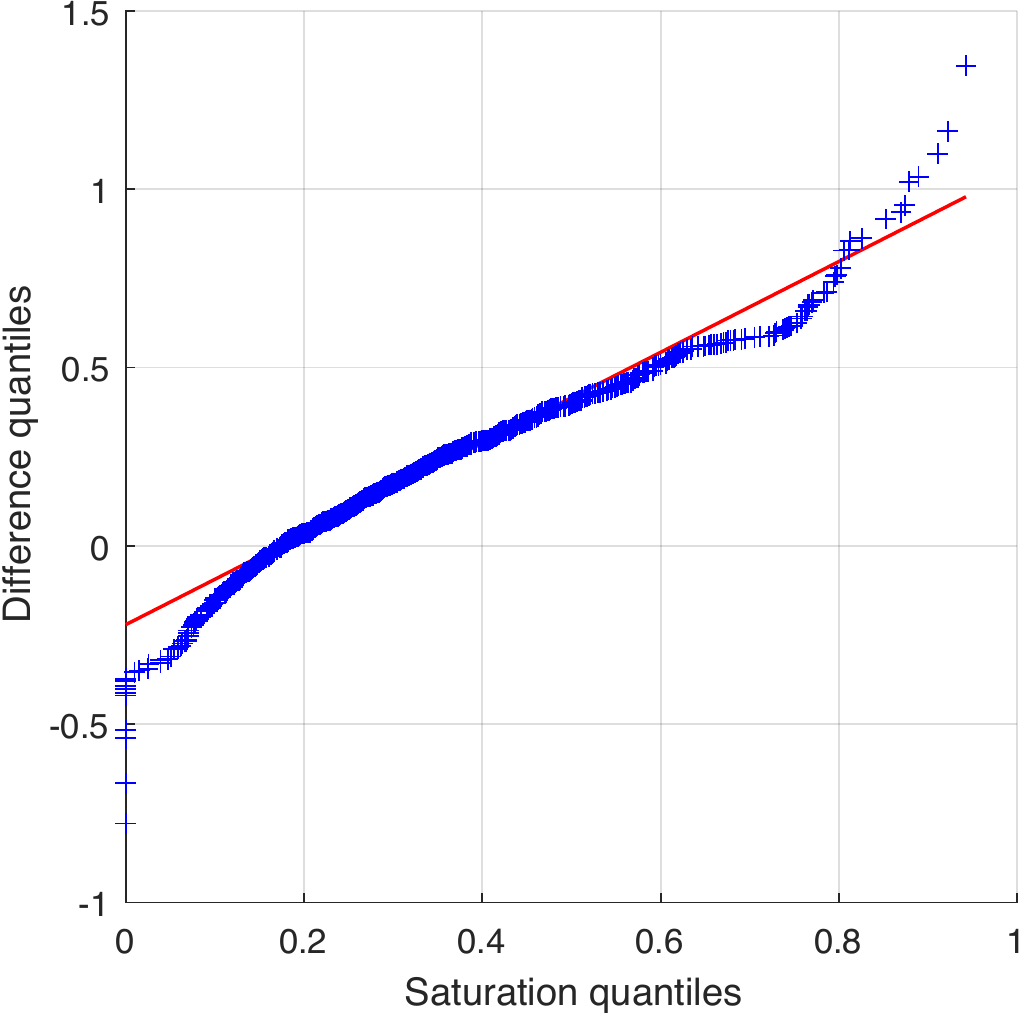}\label{fig:qqplot}
}
\subfigure[$A$ v.s. $D$]{
\includegraphics[height=2.7cm]{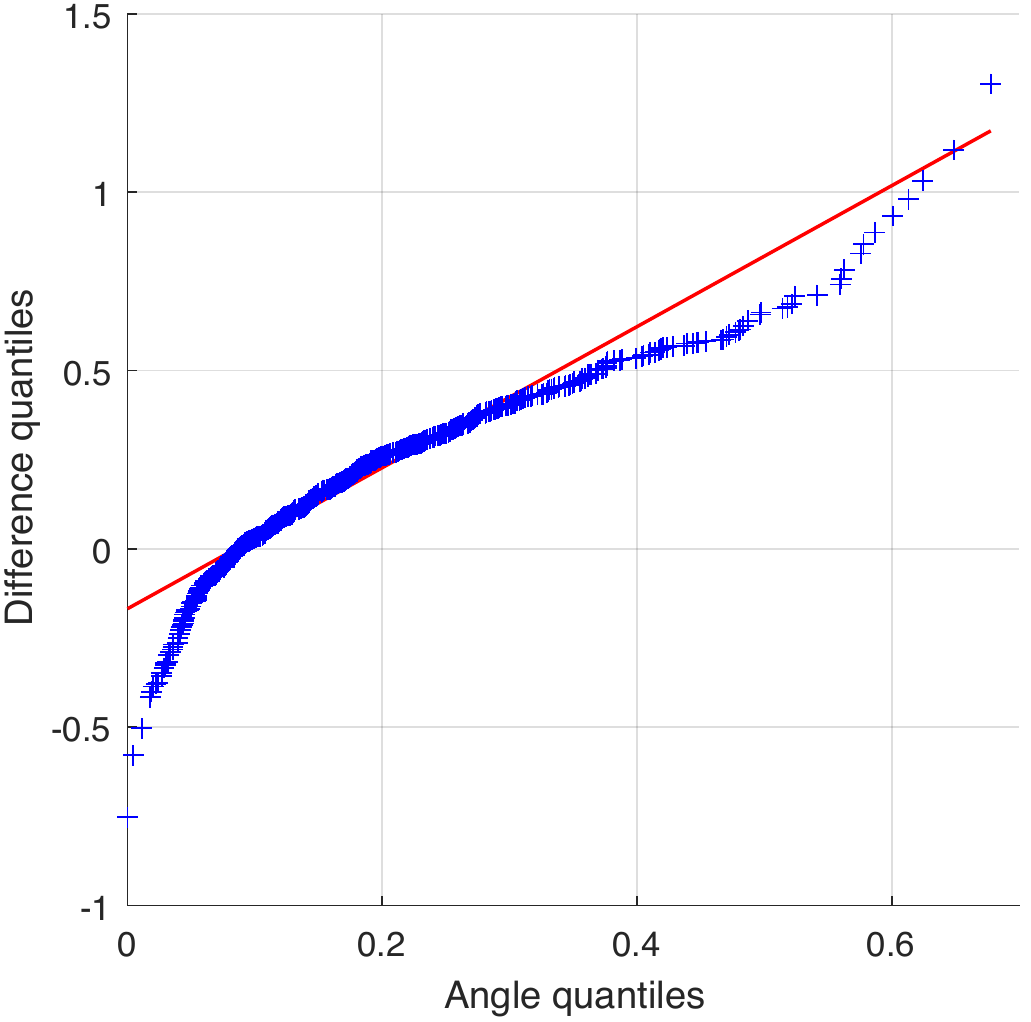}\label{fig:qqplot_angle}
}
\caption{In the denoising task, QCNN shows at least 0.5dB higher PSNR than CNN for images in (a). For images in (b), CNN offers better result.
(c) The quantile-quantile plot of saturation versus PSNR difference.
(d) The quantile-quantile plot of average angle between color vectors and gray axis versus PSNR difference.}
\label{fig:Qbetter}
\end{figure}

\subsection{Discussion on advantages and limitations}
As aforementioned, our QCNN is motivated for color image representation.
When it comes to the images with little variety of colors, our QCNN degrades to a model similar to real-valued CNN,\footnote{As we mentioned in section~\ref{ssec:qcl}, for grayscale images, QCNNs perform exactly the same as real-valued CNNs with same number of filters.} and thus, obtains just comparable or slightly worse results in the denoising task, which is confirmed on the COCO subset.

In particular, according to the results shown above, we can find that the superiority of our QCNN on the COCO subset is not so significant as that on the 102 Oxford flower data set.
To further analysis this phenomenon, we pick up those COCO images for which our QCNN shows great advantage as well as those for which our QCNN shows no advantage in the denoising task, and compare them visually in Fig.~\ref{fig:Qbetter}.
We can find that the images on which our QCNN shows better performance are often  colorful, while images where our QCNN is inferior to the real-valued CNN are close to grayscale images.

To further investigate QCNN's advantages, we use two metrics as quantitative descriptions of ``colorful images''.
The first metric is the mean saturation of color image, denoted as $S$.
For an image, a low $S$ indicates that this image is similar to a grayscale image, while a high $S$ value implies this image is with high color saturation ($i.e.$ many colorful parts).
The second metric is the averaged angle between the pixel (color vector) of color image and grayscale axis, denoted as $A$.
For an image, the larger the averaged angle is, the colorful the image is.
We show the quantile-quantile plots of these two metrics with respect to the difference between PSNR value of real-valued CNN and that of our QCNN (denoted as $D$) in Fig.~\ref{fig:qqplot} and Fig.~\ref{fig:qqplot_angle}, respectively.
We can find that both $S$ and $A$ are correlated with $D$ positively.
It means that our QCNN can show its dominant advantages over real-valued CNNs when the target images are colorful.
Otherwise, its performance is almost the same with that of real-valued CNNs.

\section{Conclusions and Future Work}
In this paper, we introduce QCNN, a quaternion-based neural network, which obtains better performance on both color image classification and color image denoising than traditional real-valued CNNs do.
A novel quaternion convolution operation is defined to represent color information in a more structural way.
A series of quaternion-based layers are designed with good compatibility to existing real-valued networks and reasonable computational complexity.
In summary, the proposed model is a valuable extension of neural network model in other number fields.
In the future, we plan to explore more efficient algorithms for the learning of QCNNs.
For example, as we mentioned in section~\ref{ssec:back}, for QCNNs their backpropagation of gradients can be represented by reverse rotations of color vectors with respect to the forward propagation of inputs.
Such a property provides us a chance to reduce the computation of the backpropagation given the intermediate information of forward propagation and  accelerate the learning of QCNNs accordingly.
Additionally, we will extend our QCNN model to large-scale data and more applications.

\section*{Acknowledgment}
This work was supported in part by National Science Foundation of China (61671298, U1611461, 61502301, 61521062), STCSM (17511105400, 17511105402, 18DZ2270700),
China's Thousand Youth Talents Plan, the 111 project B07022, the MoE Key Lab of Artificial Intelligence, AI Institute of Shanghai Jiao Tong University, and the SJTU-UCLA Joint Center for Machine Perception and Inference.
The corresponding author of this paper is Yi Xu (xuyi@sjtu.edu.cn).

%
%

%


\bibliographystyle{splncs04}
\bibliography{egbib}
\end{document}